%% file: main.tex
\documentclass[10pt, conference]{IEEEtran}
\usepackage{cite}
\usepackage{amsmath}
\usepackage{amssymb}
\usepackage{bm}
\usepackage{textcase}
\usepackage{acronym}
\usepackage{graphicx}
\usepackage{hyperref}
\usepackage{multirow}
\usepackage{siunitx}
\usepackage{tikz}
\newif\iftikzexternalizeon
\tikzexternalizeonfalse
\iftikzexternalizeon
\usetikzlibrary{external}
\fi

\usepackage{pgfplots}
\pgfplotsset{compat=1.18}
\iftikzexternalizeon
\usepgfplotslibrary{external}
\fi

\usetikzlibrary{arrows.meta,positioning,shapes.geometric,calc,matrix,fit,backgrounds,decorations.pathmorphing}

\acrodef{ICP}{iterative closest point}
\acrodef{GICP}{generalized iterative closest point}
\acrodef{PCM}{pairwise consistency maximization}
\acrodef{GNSS}{global navigation satellite system}
\acrodef{RTK}{real-time kinematic}
\acrodef{SLAM}{simultaneous localization and mapping}
\acrodef{LiDAR}{light detection and ranging}
\acrodef{NDT}{normal distributions transform}
\acrodef{RANSAC}{random sample consensus}
\acrodef{GNC}{graduated non-convexity}
\acrodef{RPE}{relative position error}
\acrodef{RRE}{relative rotation error}
\acrodef{KITTI}{KITTI}

\title{Graph Theoretical Outlier Rejection for 4D Radar Registration in Feature-Poor Environments}
\author{\IEEEauthorblockN{Georg Dorndorf} \IEEEauthorblockA{xtonomy \\ georg.dorndorf@xtonomy.ai} \and \IEEEauthorblockN{Daniel Adolfsson}  \IEEEauthorblockA{xtonomy \\ daniel.adolfsson@xtonomy.ai}  \and \IEEEauthorblockN{Masrur Doostdar}  \IEEEauthorblockA{xtonomy \\ masrur.doostdar@xtonomy.ai} }

\begin{document}
\maketitle

\begin{abstract}
Automotive 4D imaging radar is well suited for operation in dusty and low-visibility environments, but scan registration remains challenging due to scan sparsity and spurious detections caused by noise and multipath reflections.
This difficulty is compounded in feature-poor open-pit mines, where the lack of distinctive landmarks reduces correspondence reliability.

We integrate graph-based \ac{PCM} as an outlier rejection step within the \ac{ICP} loop.
We propose a radar-adapted pairwise distance-invariant scoring function for graph-based \ac{PCM} that incorporates anisotropic, per-detection uncertainty derived from a radar measurement model.

The consistency maximization problem is approximated with a greedy heuristic that finds a large clique in the pairwise consistency graph.
The refined correspondence set improves robustness when the initial association set is heavily contaminated.

We evaluate a standard Euclidean distance residual and our uncertainty-aware residual on an open-pit mine dataset collected with a 4D imaging radar. Compared to the \ac{GICP} baseline without \ac{PCM}, our method reduces segment \ac{RPE} by 29.6\% on \SI{1}{\meter} segments and by up to 55\% on \SI{100}{\meter} segments.

The presented method is intended for integration into localization pipelines and is suitable for online use due to the greedy heuristic in graph-based \ac{PCM}.
\end{abstract}
\section{Introduction}
Reliable localization is critical for autonomous operation in open-pit mines and other large, feature-poor environments.
Automotive 4D imaging radar is particularly attractive in this setting due to its robustness to adverse environmental effects such as dust.

In practice, autonomous mining systems often rely on \ac{GNSS}, but coverage can be intermittently degraded or denied due to terrain occlusions, infrastructure, or atmospheric effects.
During such outages, radar-based scan registration can provide relative motion estimates as part of the localization pipeline.

Automotive 4D imaging radars provide detections in range, azimuth, elevation, and Doppler; in this work, we focus on the 3D spatial measurements.
Compared to \ac{LiDAR}, the resulting point clouds are much sparser and are frequently affected by spurious detections and outliers caused by noise and multipath reflections, including ghost targets.
As a result, correspondence-based registration can become unstable unless outlier correspondences are filtered.
This motivates correspondence refinement and outlier rejection mechanisms that can tolerate heavily contaminated initial association sets.
Consequently, \ac{ICP}-style radar registration is typically combined with robust estimation or outlier rejection to reduce the impact of incorrect correspondences.

We construct a consistency graph over candidate correspondences and select a mutually consistent subset using graph-based \ac{PCM}.
To reflect radar sensing, we introduce a pairwise consistency test that accounts for anisotropic per-detection uncertainty.
For efficiency, the consistency maximization is approximated with a greedy large-clique heuristic.
The refined correspondences are then used inside the \ac{ICP} iterations to improve robustness in feature-poor scenes.

The core contributions are:
\begin{itemize}
    \item an uncertainty-aware pairwise distance invariant for 4D radar correspondences,
    \item integration of graph-based \ac{PCM} as an inlier selection step within radar \ac{ICP},
    \item evaluation on an open-pit mine dataset recorded with a 4D imaging radar.
\end{itemize}

\section{Related Work}
Robust scan matching with automotive radar has been studied both as a component in radar odometry/\ac{SLAM} and for map-based localization. Early radar \ac{SLAM} systems have used \ac{ICP} variants on radar detections \cite{holder_real-time_2019}, while recent 4D imaging radar pipelines combine scan matching with Doppler-based motion estimation and filtering to improve robustness in sparse and noisy point clouds \cite{li_4d_2023,zhang_4dradarslam_2023,kubelka_we_2024}. For registration itself, choices include distribution-based alignment such as \ac{NDT} \cite{biber_normal_2003} and uncertainty-aware variants of \ac{GICP} that incorporate per-point measurement uncertainty \cite{zhang_4dradarslam_2023,xu_incorporating_2025}.

Outlier rejection remains a key difficulty due to spurious detections from noise and multipath reflections, as well as low scan overlap. Robust estimation techniques such as \ac{RANSAC} and \ac{GNC} are used in radar odometry and registration pipelines \cite{doer_ekf_2020,lim_orora_2023,stironja_rave_2024}. Recent radar odometry methods also explicitly leverage anisotropic measurement characteristics, e.g., via polar measurement models \cite{retan_radar_2022} or anisotropy-aware estimation components \cite{lim_orora_2023}.

In parallel, \ac{PCM} has emerged as a correspondence-refinement strategy in point cloud registration \cite{mangelson_pairwise_2018}. Graph-based \ac{PCM} methods build a compatibility graph over putative correspondences and seek a mutually consistent inlier set, commonly via maximum clique or relaxations thereof, as in ROBIN \cite{shi_robin_2021} and CLIPPER/CLIPPER+ \cite{lusk_clipper_2021,fathian_clipper_2024}. Pairwise-consistency maximization has also been used in 2D-radar \ac{SLAM}. For example, RadarSLAM applies a pairwise distance-consistency constraint and selects a mutually consistent inlier set via a maximum clique formulation for 2D radar keypoint matches \cite{hong_radarslam_2020}.

This work adapts graph-based \ac{PCM} to 4D imaging radar registration by defining an uncertainty-aware pairwise invariant that accounts for anisotropic per-detection uncertainty \cite{retan_radar_2022,lim_orora_2023}. The refined correspondence set is then used to robustify \ac{ICP} in feature-poor environments.

\section{Methodology}
We consider the problem of aligning two point clouds by estimating a rigid-body transform \(\mathbf{T}\in SE(3)\) that maps the source point set \(S\) into the target point set \(R\). We employ graph-based \ac{PCM} to refine radar detection correspondences prior to each transform update.
Specifically, within each iteration we generate putative correspondences, build a pairwise-consistency graph using either a Euclidean or uncertainty-aware invariant, select a mutually consistent subset via a greedy clique heuristic, and estimate the transform using an \ac{ICP}/\ac{GICP}-style update.

Let \(S = \{\mathbf{p}_i\}\) and \(R = \{\mathbf{r}_j\}\) be two sets of points, representing the source and the target point clouds, respectively. Given a candidate association set \(\mathcal{A} \subset S \times R\), where \(a=(\mathbf{p},\mathbf{r})\in \mathcal{A}\), the transform is estimated by minimizing a correspondence-based registration objective of the form
\begin{align}
    \min_{\mathbf{T}\in SE(3)} \sum_{(\mathbf{p},\mathbf{r})\in \mathcal{I}} \rho\!\left(e\!\left(\mathbf{T},\mathbf{p},\mathbf{r}\right)\right),
\end{align}
where \(\mathcal{I}\subseteq\mathcal{A}\) denotes the inlier correspondence set, \(\rho(\cdot)\) is an optional robust loss, and \(e(\cdot)\) is a geometric residual (e.g., point-to-point, point-to-plane, or \ac{GICP}).

\subsection{Putative Correspondences}
We construct a set of putative correspondences \(\mathcal{A}\subset S\times R\) to initialize registration. In our experiments, we use a simple 1-nearest-neighbor strategy: for each source point \(\mathbf{p}_i\in S\) we associate the closest target point in Euclidean distance,
\begin{align}
    \mathcal{A} = \{(\mathbf{p}_i, \mathbf{r}_{j(i)})\mid j(i)=\arg\min_j \lVert \mathbf{p}_i-\mathbf{r}_j\rVert_2\}.
\end{align}
This yields at most one candidate match per source point and keeps the correspondence set size tractable for graph-based \ac{PCM}. Note that \(\mathcal{A}\) may contain repeated target points. We do not evaluate multi-hypothesis or uncertainty-driven correspondence generation (e.g., for scan-to-map registration) in this work.

\subsection{Consistency Graph Construction}\label{subsec:consistency-graph}
We build a consistency graph \(G=(V,E)\) over the putative correspondences, with \(V=\mathcal{A}\). The pairwise consistency \emph{score} for two associations \(a=(\mathbf{p},\mathbf{r}),\,b=(\mathbf{q},\mathbf{s}) \in \mathcal{A}\) is denoted \(s(a,b)\), where smaller values indicate higher consistency. We consider two variants: a raw distance-difference score \(s_{\text{raw}}(a,b)=\big\lvert\lVert \mathbf{p}-\mathbf{q}\rVert_2-\lVert \mathbf{r}-\mathbf{s}\rVert_2\big\rvert\) and an uncertainty-normalized score \(s_{\text{norm}}(a,b)=v^2/\sigma_v^2\) defined in Section~\ref{subsec:radar-invariant}. The raw score is thresholded with $\tau$ (\si{\meter}) and the normalized score with $\alpha$ (unitless).

For a threshold \(\gamma > 0\) (either \(\gamma=\tau\) for $s_{\text{raw}}$ or \(\gamma=\alpha\) for $s_{\text{norm}}$), the edges between the vertices are:
\begin{align}
    E=\{(a,b) \mid &\ a=(\mathbf{p},\mathbf{r})\in V,\ b=(\mathbf{q},\mathbf{s})\in V,\ s_{\bullet}(a,b) < \gamma \nonumber \\
    &\land \mathbf{p} \neq \mathbf{q} \land \mathbf{r} \neq \mathbf{s} \},
\end{align}
where $s_{\bullet}\in\{s_{\text{raw}},s_{\text{norm}}\}$ denotes the chosen score.
The additional constraints \(\mathbf{p}\neq\mathbf{q}\) and \(\mathbf{r}\neq\mathbf{s}\) prevent selecting two correspondences that share the same source or target point.

\begin{figure}[t]
    \centering
    \resizebox{\columnwidth}{!}{\input{consistency-graph}}
    \caption{Toy example illustrating (left) putative associations between two scans and (right) the induced consistency graph used by \ac{PCM}.}
    \label{fig:consistency_graph_toy}
\end{figure}

A maximum clique in this graph is the largest set of mutually consistent associations.

\subsection{Radar Uncertainty-Aware Invariant} \label{subsec:radar-invariant}
\subsubsection{Radar Measurement Model}
Radar detections exhibit anisotropic uncertainty that is commonly expressed by the variances of the components of the points in spherical coordinates.
A 4D radar detection provides measurements in range \(r\), azimuth \(\theta\), elevation \(\phi\), and Doppler \(d\). We limit ourselves to the first three.
The measurement uncertainties are assumed independent, resulting in a diagonal covariance matrix \(\mathbf{\Sigma}_\text{d} \in \mathbb{S}^3_{+}\) (symmetric positive semidefinite):
\begin{align}
    \mathbf{\Sigma}_\text{d} = \begin{bmatrix}
        \sigma_r^2 &0&0\\
        0&\sigma_\theta^2 &0\\
        0&0&\sigma_\phi^2
    \end{bmatrix}
\end{align}
To use this uncertainty in Cartesian space, the spherical-to-Cartesian point transformation is linearized using its Jacobian \(\mathbf{J}_{\text{sc}}\), evaluated at the measured \((r,\theta,\phi)\).
\begin{align}
    \mathbf{\Sigma}_\text{e} = \mathbf{J}_{\text{sc}} \, \mathbf{\Sigma}_\text{d} \, \mathbf{J}_{\text{sc}}^\top
\end{align}
where \(\mathbf{\Sigma}_\text{e}\) represents the point-wise uncertainty in Euclidean space, and we write \(\mathbf{\Sigma}_{\mathbf{p}}\) for the instance associated with point \(\mathbf{p}\).

\subsubsection{Uncertainty-normalized consistency score}
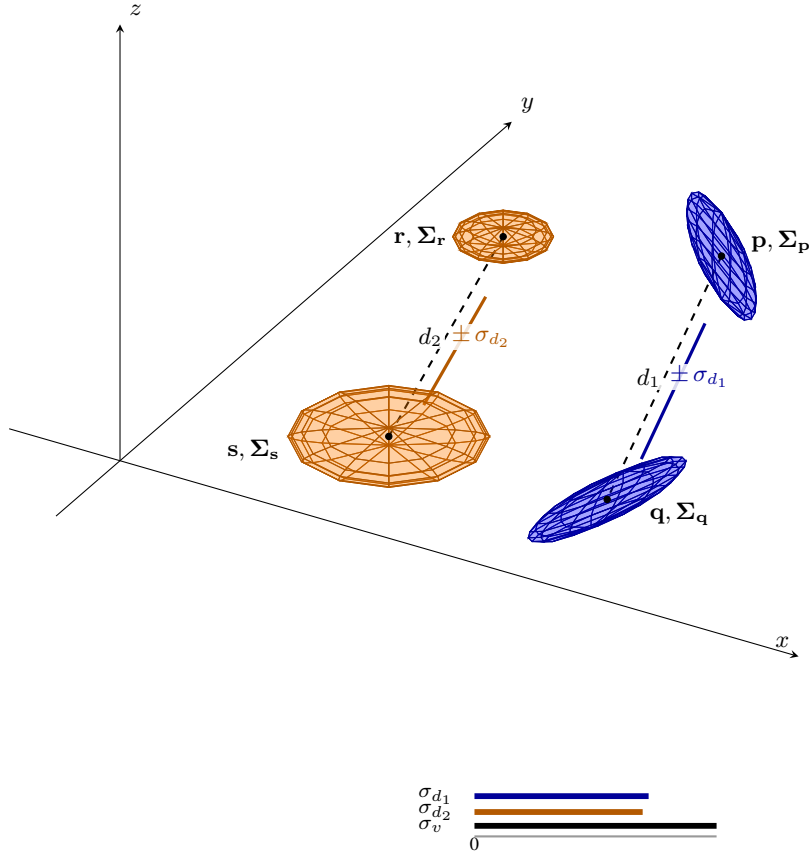
\begin{figure*}[t]
    \centering
    \input{radar-pcm-invariant}
    \caption{Visualization of the pairwise distance invariant for two correspondences $(\mathbf{p},\mathbf{r})$ and $(\mathbf{q},\mathbf{s})$. Shown are 1$\sigma$ uncertainty ellipsoids and their projected 1$\sigma$ uncertainties along the distance directions used in $\sigma_v^2$.}
    \label{fig:invariant_projection}
\end{figure*}

Given two associations \((\mathbf{p},\mathbf{r}),(\mathbf{q},\mathbf{s}) \in \mathcal{A}\), together with the point-wise uncertainties \(\mathbf{\Sigma}_{\mathbf{p}}, \mathbf{\Sigma}_{\mathbf{q}},\mathbf{\Sigma}_{\mathbf{r}},\mathbf{\Sigma}_{\mathbf{s}}\), we define the pairwise distance-difference residual:
\begin{align}
v(\mathbf{p}, \mathbf{q}, \mathbf{r},\mathbf{s}) = \lVert \mathbf{p}-\mathbf{q}\rVert_2 - \lVert \mathbf{r}-\mathbf{s}\rVert_2.
\end{align}
We then evaluate an uncertainty-normalized (Mahalanobis-style) score relative to zero, which serves as our normalized consistency score:
\begin{align}
    s_{\text{norm}} &= \frac{v^2}{\sigma_v^2}
\end{align}
To find the combined variance \(\sigma_v^2\), we linearize \(v\) around the operating points \(\mathbf{p}_0, \mathbf{q}_0, \mathbf{r}_0, \mathbf{s}_0\):
\begin{align}
   v \approx v_0 &+ \left.\nabla_\mathbf{p} v\right|_0^\top (\mathbf{p} - \mathbf{p}_0) + \left.\nabla_\mathbf{q} v\right|_0^\top (\mathbf{q} - \mathbf{q}_0) \nonumber \\
    &+ \left.\nabla_\mathbf{r} v\right|_0^\top (\mathbf{r} - \mathbf{r}_0) + \left.\nabla_\mathbf{s} v\right|_0^\top (\mathbf{s} - \mathbf{s}_0)
\end{align}
Assuming \(\mathbf{p}_0 \neq \mathbf{q}_0\) and \(\mathbf{r}_0 \neq \mathbf{s}_0\), the gradients evaluated at the operating points represent the normalized directional vectors between the point pairs:
\begin{align}
     \left.\nabla_\mathbf{p} v\right|_0 &= \frac{\mathbf{p}_0-\mathbf{q}_0}{\lVert \mathbf{p}_0-\mathbf{q}_0 \rVert_2}, & \left.\nabla_\mathbf{q} v\right|_0 &= -\left.\nabla_\mathbf{p} v\right|_0, \\
    \left.\nabla_\mathbf{r} v\right|_0 &= -\frac{\mathbf{r}_0-\mathbf{s}_0}{\lVert \mathbf{r}_0-\mathbf{s}_0 \rVert_2}, & \left.\nabla_\mathbf{s} v\right|_0 &= -\left.\nabla_\mathbf{r} v\right|_0.
\end{align}
Here, for \(\mathbf{x} \in \{\mathbf{p},\mathbf{q},\mathbf{r},\mathbf{s}\}\), we define the Jacobian row \(\mathbf{J}_{\mathbf{x}} := \left.\nabla_{\mathbf{x}} v\right|_0^\top\). Assuming independent point uncertainties, applying the linear error propagation \(\sigma_v^2 = \sum_i \mathbf{J}_i\, \mathbf{\Sigma}_i\, \mathbf{J}_i^\top\) yields:
\begin{align}
      \sigma_v^2 = &\frac{(\mathbf{p}-\mathbf{q})^\top (\mathbf{\Sigma}_{\mathbf{p}} + \mathbf{\Sigma}_{\mathbf{q}}) (\mathbf{p}-\mathbf{q})}{\lVert \mathbf{p}-\mathbf{q} \rVert_2^2} \nonumber \\
    &+ \frac{(\mathbf{r}-\mathbf{s})^\top (\mathbf{\Sigma}_{\mathbf{r}} + \mathbf{\Sigma}_{\mathbf{s}}) (\mathbf{r}-\mathbf{s})}{\lVert \mathbf{r}-\mathbf{s} \rVert_2^2}
\end{align}
This projects the 3D point uncertainty ellipsoids down into the space of 1D distance residuals. Fig.~\ref{fig:invariant_projection} illustrates the four-point configuration and the projection of the 3D uncertainties onto the distance directions.
The threshold \(\alpha\) can then be set from a chi-square quantile with 1 degree of freedom.

\subsection{Maximum Clique Heuristic}
Graph-based \ac{PCM} seeks a large mutually consistent correspondence set, which can be posed as a maximum clique problem on the consistency graph. Let \(n:=|V|=|\mathcal{A}|\) and \(m:=|E|\). In our setting, \(\mathcal{A}\) is generated by 1-nearest-neighbor matching, hence \(n\le |S|\). Since \ac{PCM} is invoked at every \ac{ICP} iteration, exact maximum clique solvers are typically too costly.

We use a greedy heuristic based on a smallest-last (minimum-degree) vertex ordering \cite{matula_smallest-last_1983,walteros_why_2020}. The ordering can be computed in \(\mathcal{O}(n+m)\) time \cite{matula_smallest-last_1983}, after which we traverse the ordering in reverse and greedily add a vertex if it is adjacent to all vertices in the current clique. With adjacency lists, the clique construction step has worst-case time \(\mathcal{O}(n\,|C|)\subseteq\mathcal{O}(n^2)\), where \(|C|\) is the returned clique size.

Note that building the consistency graph itself requires evaluating the pairwise cost \(s(a,b)\) for candidate pairs \((a,b)\), which is \(\mathcal{O}(n^2)\) in the worst case and dominates the overall worst-case execution time when the graph is dense.

\subsection{\texorpdfstring{\ac{ICP}}{ICP} with pairwise consistency maximization}
\begin{figure*}[t]
    \centering
    \resizebox{0.92\textwidth}{!}{\input{icp-pcm}}
    \caption{Integration of graph-based \ac{PCM} into the \ac{ICP} loop for correspondence refinement.}
    \label{fig:icp_pcm}
\end{figure*}
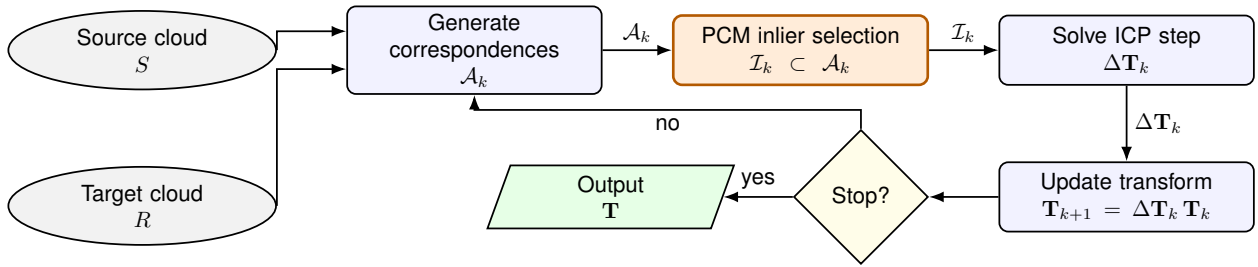

We integrate \ac{PCM} as an inlier selection step inside a standard \ac{ICP} loop (Fig.~\ref{fig:icp_pcm}). Given a candidate correspondence set $\mathcal{A}_k$ at iteration $k$, we select a mutually consistent inlier subset $\mathcal{I}_k \subset \mathcal{A}_k$ using the consistency graph construction in Section~\ref{subsec:consistency-graph} and the radar uncertainty-aware pairwise cost in Section~\ref{subsec:radar-invariant}. We then estimate an incremental rigid transform $\Delta\mathbf{T}_k$ from $\mathcal{I}_k$ using a conventional ICP objective, point-to-point \cite{besl_method_1992}, point-to-plane \cite{chen_object_1992} or Generalized ICP \cite{segal_generalized-icp_2010} and update $\mathbf{T}_{k+1}=\Delta\mathbf{T}_k\,\mathbf{T}_k$ until convergence.

\section{Experiments}\label{sec:experiments}
We evaluate scan-to-scan registration on radar data collected in a feature-poor open-pit mine in the Austrian Alps.
A Continental ARS548 4D imaging radar is mounted front-facing on the bumper of a Bell B30E articulated dump truck and operated at \SI{20}{\hertz}.
Each scan corresponds to a single radar frame.

\subsection{Experimental Setup}
We register consecutive scans with an identity initial guess \(\mathbf{T}_0=\mathbf{I}\) and integrate the resulting relative transforms into a radar-only odometry trajectory, which provides a stress test for scan registration in sparse, outlier-contaminated radar point clouds.
Putative correspondences are generated by 1-nearest-neighbor matching in Euclidean space with a maximum association distance of \SI{10}{\meter}.

The dataset contains a sequence recorded in an open-pit mine; the driven trajectory spans two mine levels connected via a ramp and has a total length of \SI{4705}{\meter} over \SI{21}{\minute}, resulting in \num{25155} radar scans.
This corresponds to an average inter-frame translation of $\approx\SI{0.19}{\meter}$, making an identity initialization reasonable.
No dynamic objects are present in the recording, and no pre-filtering is applied to the radar detections before registration.

For radar per-point uncertainties, we use the manufacturer-provided accuracies to form detection covariances; no environment-specific calibration is performed.

Ground truth is provided by \ac{RTK}-fix \ac{GNSS}; its accuracy in the mine environment is not independently validated and may include decimeter-level errors.

We report the following registration configurations:
\begin{itemize}
    \item \textbf{\ac{ICP} (Pt2Pt):} 1-nearest-neighbor correspondences with a point-to-point objective.
    \item \textbf{\ac{GICP}:} per-point covariances derived from radar detections.
    \item \textbf{\ac{ICP}/\ac{GICP} + \ac{PCM} (raw score):} \ac{PCM} with the raw distance-difference score $s_{\text{raw}}$ and an edge threshold $\tau$ (\si{\meter}).
    \item \textbf{\ac{ICP}/\ac{GICP} + \ac{PCM} (uncertainty-normalized score):} \ac{PCM} with the uncertainty-normalized score $s_{\text{norm}}$ (Section~\ref{subsec:radar-invariant}) and the unitless edge threshold $\alpha$.
\end{itemize}
We sweep $\tau\in\{0.25,0.50,1.00,3.86\}\,\si{\meter}$ and $\alpha\in\{0.016,0.25,1.00,2.706,3.86,5.00,6.63,10.83\}$ for \textbf{\ac{ICP} (Pt2Pt) + \ac{PCM}} and select thresholds for all \ac{PCM}-augmented methods based on this sweep.
We report, for each segment length, the mean $\pm$ one standard deviation across all evaluation segments (Table~\ref{tab:results_mine}).

\subsection{Metrics}
We report segment-level \ac{RPE} over fixed path-length segments of \SI{1}{\meter}, \SI{5}{\meter}, \SI{10}{\meter}, \SI{20}{\meter}, \SI{50}{\meter}, and \SI{100}{\meter} in translation and \ac{RRE} in rotation, computed from the estimated (integrated) radar-only trajectory and the \ac{RTK}-fix \ac{GNSS} ground truth.
Additionally, we report \acs{KITTI} odometry-style metrics (average translational drift $t_{\text{rel}}$ [\si{\percent}] and rotational drift $r_{\text{rel}}$ [\si{\degree\per\SI{100}{\meter}}]), computed by averaging relative drift over trajectory segments as in the \acs{KITTI} odometry benchmark.
To mitigate the impact of local ground-truth inaccuracies on single scan-to-scan alignment, we report \ac{RPE} over fixed path-length segments rather than per-scan-pair errors.
The uncertainty-aware invariant depends on a calibrated measurement uncertainty model; using manufacturer-provided, fixed spherical accuracies is a simplification, and calibrating these covariances for the mine setting is left as future work.

\begin{table*}[t]
\centering
\caption{Trajectory errors for scan-to-scan registration integrated into a radar-only trajectory (for each segment length: mean $\pm$ one standard deviation across all segments). Baselines are the methods without \ac{PCM}; our approach corresponds to the \ac{PCM}-augmented methods. For the raw (Euclidean) score, the edge threshold is $\tau$ (\si{\meter}); for the uncertainty-normalized (Mahalanobis-style) score it is the unitless cutoff $\alpha$. Best (lowest) $t_{\text{rel}}$ within each score family and best (lowest) RPE/RRE per segment length are shown in bold.}
\label{tab:results_mine}
\setlength{\tabcolsep}{3pt}
\renewcommand{\arraystretch}{1.10}
\resizebox{\textwidth}{!}{%
\begin{tabular}{lcccccccccccccccccc}
\hline
\multirow{2}{*}{Method} & \multirow{2}{*}{PCM} & \multirow{2}{*}{Invariant} & \multirow{2}{*}{Thr.} & \multicolumn{2}{c}{KITTI} & \multicolumn{12}{c}{Segment metrics}\\
\cline{5-6}\cline{7-18}
 &  &  &  & $t_{\text{rel}}$ [\si{\percent}] & $r_{\text{rel}}$ [\si{\degree\per\SI{100}{\meter}}] & \multicolumn{2}{c}{\SI{1}{\meter}} & \multicolumn{2}{c}{\SI{5}{\meter}} & \multicolumn{2}{c}{\SI{10}{\meter}} & \multicolumn{2}{c}{\SI{20}{\meter}} & \multicolumn{2}{c}{\SI{50}{\meter}} & \multicolumn{2}{c}{\SI{100}{\meter}}\\
\cline{7-18}
 &  &  &  &  &  & RPE [\si{\meter}] & RRE [\si{\degree}] & RPE [\si{\meter}] & RRE [\si{\degree}] & RPE [\si{\meter}] & RRE [\si{\degree}] & RPE [\si{\meter}] & RRE [\si{\degree}] & RPE [\si{\meter}] & RRE [\si{\degree}] & RPE [\si{\meter}] & RRE [\si{\degree}]\\
\hline
GICP & No & --- & --- & $19.68$ & $33.02$ & $0.348\pm0.209$ & $0.696\pm0.551$ & $1.047\pm0.565$ & $1.805\pm1.598$ & $1.841\pm0.949$ & $3.077\pm2.802$ & $3.256\pm1.544$ & $5.318\pm4.776$ & $7.044\pm2.885$ & $10.005\pm5.675$ & $13.597\pm6.135$ & $15.042\pm6.608$\\
ICP (Pt2Pt) & No & --- & --- & $43.33$ & $45.83$ & $0.768\pm0.380$ & $0.995\pm0.785$ & $2.235\pm0.956$ & $2.448\pm1.763$ & $3.976\pm1.558$ & $4.005\pm2.850$ & $7.294\pm2.690$ & $6.871\pm4.851$ & $16.466\pm5.279$ & $14.115\pm8.116$ & $29.281\pm9.488$ & $23.874\pm11.461$\\
\hline
GICP + PCM & Yes & Euclidean & 0.25 & $\bm{11.13}$ & $31.10$ & $\bm{0.245}\pm0.182$ & $0.815\pm0.766$ & $\bm{0.613}\pm0.486$ & $1.817\pm1.866$ & $0.980\pm0.829$ & $2.685\pm2.953$ & $\bm{1.586}\pm1.361$ & $4.165\pm4.409$ & $\bm{3.088}\pm1.662$ & $6.141\pm5.037$ & $\bm{6.122}\pm4.460$ & $8.776\pm6.961$\\
GICP + PCM & Yes & Mahalanobis & 5.00 & $\bm{13.34}$ & $23.68$ & $0.294\pm0.350$ & $\bm{0.613}\pm0.495$ & $0.751\pm0.723$ & $\bm{1.341}\pm1.545$ & $1.196\pm1.046$ & $\bm{2.068}\pm2.701$ & $1.962\pm1.666$ & $\bm{3.276}\pm4.780$ & $3.768\pm2.099$ & $\bm{4.616}\pm4.330$ & $6.292\pm3.365$ & $\bm{7.736}\pm7.740$\\
\hline
\multirow{4}{*}{ICP (Pt2Pt) + PCM} & \multirow{4}{*}{Yes} & \multirow{4}{*}{Euclidean} & 0.25 & $11.39$ & $35.02$ & $0.251\pm0.182$ & $0.919\pm0.814$ & $0.623\pm0.456$ & $2.075\pm1.938$ & $\bm{0.979}\pm0.775$ & $3.018\pm3.094$ & $1.626\pm1.388$ & $4.635\pm4.942$ & $3.185\pm1.825$ & $6.725\pm5.087$ & $6.435\pm3.599$ & $9.871\pm7.487$\\
 &  &  & 0.50 & $12.35$ & $30.07$ & $0.246\pm0.153$ & $0.768\pm0.624$ & $0.676\pm0.474$ & $1.745\pm1.785$ & $1.144\pm0.842$ & $2.666\pm3.062$ & $1.954\pm1.562$ & $4.016\pm5.021$ & $4.146\pm2.258$ & $6.619\pm6.778$ & $6.506\pm3.111$ & $8.731\pm6.324$\\
 &  &  & 1.00 & $16.84$ & $29.50$ & $0.297\pm0.176$ & $0.736\pm0.640$ & $0.919\pm0.586$ & $1.700\pm1.856$ & $1.620\pm1.032$ & $2.598\pm3.246$ & $2.906\pm1.761$ & $4.006\pm5.110$ & $6.091\pm3.274$ & $6.464\pm5.095$ & $10.116\pm6.004$ & $10.469\pm8.935$\\
 &  &  & 3.86 & $101.56$ & $55.89$ & $1.269\pm0.417$ & $1.010\pm0.780$ & $5.351\pm1.393$ & $3.059\pm2.382$ & $10.379\pm2.607$ & $5.351\pm3.962$ & $20.125\pm4.844$ & $9.431\pm6.792$ & $45.890\pm12.259$ & $19.841\pm12.293$ & $79.203\pm24.272$ & $32.774\pm18.854$\\
\hline
\multirow{8}{*}{ICP (Pt2Pt) + PCM} & \multirow{8}{*}{Yes} & \multirow{8}{*}{Mahalanobis} & 0.016 & $564.78$ & $585.46$ & $11.801\pm24.568$ & $15.690\pm34.042$ & $42.528\pm51.874$ & $37.226\pm48.853$ & $71.041\pm66.020$ & $48.859\pm50.790$ & $87.831\pm69.453$ & $80.230\pm57.138$ & $71.619\pm18.402$ & $108.400\pm45.655$ & $65.272\pm24.986$ & $92.733\pm45.427$\\
 &  &  & 0.25 & $26.59$ & $40.31$ & $0.605\pm1.380$ & $1.006\pm1.736$ & $1.586\pm2.711$ & $2.296\pm3.857$ & $2.440\pm3.565$ & $3.400\pm5.567$ & $3.655\pm4.811$ & $5.228\pm7.622$ & $6.609\pm6.954$ & $9.801\pm13.122$ & $11.460\pm11.841$ & $15.567\pm19.559$\\
 &  &  & 1.00 & $15.81$ & $30.18$ & $0.374\pm0.568$ & $0.799\pm0.711$ & $0.926\pm1.155$ & $1.735\pm1.815$ & $1.419\pm1.659$ & $2.639\pm3.013$ & $2.179\pm2.388$ & $4.002\pm4.762$ & $3.657\pm2.490$ & $5.769\pm4.894$ & $6.494\pm4.076$ & $8.577\pm5.647$\\
 &  &  & 2.706 & $13.88$ & $28.82$ & $0.317\pm0.354$ & $0.739\pm0.656$ & $0.780\pm0.735$ & $1.661\pm1.800$ & $1.212\pm1.042$ & $2.470\pm2.994$ & $1.988\pm1.622$ & $3.851\pm4.607$ & $3.684\pm2.144$ & $6.157\pm5.202$ & $6.519\pm3.985$ & $9.540\pm6.724$\\
 &  &  & 3.86 & $13.83$ & $28.31$ & $0.309\pm0.328$ & $0.728\pm0.630$ & $0.773\pm0.703$ & $1.634\pm1.690$ & $1.208\pm1.040$ & $2.406\pm2.852$ & $2.020\pm1.759$ & $3.729\pm4.647$ & $3.751\pm2.307$ & $6.068\pm6.492$ & $6.991\pm3.973$ & $9.571\pm7.659$\\
 &  &  & 5.00 & $13.76$ & $28.22$ & $0.307\pm0.301$ & $0.724\pm0.647$ & $0.770\pm0.661$ & $1.614\pm1.757$ & $1.220\pm0.988$ & $2.426\pm2.892$ & $1.968\pm1.454$ & $3.779\pm4.383$ & $3.676\pm1.754$ & $5.934\pm4.655$ & $7.040\pm3.606$ & $9.582\pm7.390$\\
 &  &  & 6.63 & $14.23$ & $28.08$ & $0.310\pm0.290$ & $0.712\pm0.612$ & $0.777\pm0.626$ & $1.594\pm1.671$ & $1.269\pm0.951$ & $2.461\pm2.851$ & $2.071\pm1.475$ & $3.755\pm4.589$ & $4.051\pm2.233$ & $6.067\pm5.252$ & $7.677\pm4.118$ & $9.827\pm5.890$\\
 &  &  & 10.83 & $14.97$ & $28.68$ & $0.319\pm0.266$ & $0.719\pm0.606$ & $0.846\pm0.615$ & $1.651\pm1.738$ & $1.329\pm0.872$ & $2.523\pm2.845$ & $2.231\pm1.404$ & $3.927\pm4.613$ & $4.362\pm1.911$ & $6.244\pm4.676$ & $7.886\pm3.407$ & $9.826\pm6.858$\\
\hline
\end{tabular}}
\end{table*}
\subsection{Results and Discussion}
Table~\ref{tab:results_mine} shows that incorporating \ac{PCM} consistently improves robustness over the corresponding baselines without \ac{PCM}, reducing both segment-level \ac{RPE}/\ac{RRE} and overall drift metrics.

In the table, the entries labeled ``Euclidean'' correspond to the raw score $s_{\text{raw}}$ and those labeled ``Mahalanobis'' correspond to the uncertainty-normalized score $s_{\text{norm}}$.
In particular, \ac{GICP} + \ac{PCM} achieves the best translational drift with the raw (Euclidean) score threshold $\tau=0.25$ (lowest $t_{\text{rel}}$), while the uncertainty-normalized score with $\alpha=5.0$ achieves comparable drift with lower rotational drift (compared to \ac{GICP} + \ac{PCM} with the raw score).
\ac{GICP} + \ac{PCM} with the Euclidean score and threshold $\tau=0.25$ reduces the segment \ac{RPE} by 29.6\% for \SI{1}{\meter} segments and by up to 55\% for \SI{100}{\meter} segments.

Overall, the raw score performs slightly better in translation on this dataset, which may be explained by the use of uncalibrated, manufacturer-provided uncertainty values in the uncertainty-normalized score.

For the uncertainty-aware invariant, extremely strict thresholds (e.g., very small $\alpha$) can be brittle and lead to failure, whereas a broad range of moderate $\alpha$ values yields similar drift, indicating reduced sensitivity once the acceptance region is not overly restrictive.

\section{Conclusion}
This paper presented a correspondence refinement approach for 4D radar scan registration based on graph-based \ac{PCM} with an uncertainty-aware pairwise invariant.
By selecting a mutually consistent inlier set within each \ac{ICP} iteration, the method significantly increases robustness to spurious radar detections and multipath-induced outliers in feature-poor mining environments. In our experiments, \ac{GICP} + \ac{PCM} with $\tau=0.25$ reduced segment \ac{RPE} by 29.6\% for \SI{1}{\meter} segments and by up to 55\% for \SI{100}{\meter} segments.

A limitation of the current study is the restricted set of baselines; a broader comparison to additional robust registration methods is of interest.
Furthermore, the ground-truth accuracy is not validated and is likely on the order of a few decimeters. Therefore, the reported errors are likely pessimistic.

Learning a radar uncertainty model from ground-truth data is of interest to evaluate whether the proposed adaptation can outperform the basic distance-based approach.

Future work may investigate the integration of this registration component into a full robust 4D radar localization pipeline.
Overall, integrating \ac{PCM} into \ac{ICP} provides an effective mechanism to improve radar scan registration robustness in feature-poor mine environments.
\bibliographystyle{IEEEtran}
\bibliography{IEEEabrv, references}
\end{document}

%% file: consistency-graph.tex
\begin{tikzpicture}[font=\footnotesize, node distance=1.6cm and 2.2cm]

\tikzset{
    point/.style={circle, inner sep=0pt, minimum size=5.2pt, draw=black, line width=0.5pt},
    source_point/.style={point, fill=black!15, label={[font=\scriptsize]above:$#1$}},
    target_point/.style={point, fill=white, label={[font=\scriptsize]below:$#1$}},
    association_line/.style={draw=black, line width=0.6pt, {Stealth[length=2mm]}-{Stealth[length=2mm]}},
    graph_node/.style={circle, draw=black, line width=0.6pt, minimum size=4.8mm, inner sep=1.5pt, font=\scriptsize, fill=white},
    consistent_edge/.style={draw=black, line width=0.7pt},
    clique_highlight/.style={fill=black!10, draw=black!60, line width=0.4pt, rounded corners, fill opacity=0.6, inner sep=3pt},
    cloud_container/.style={fill=black!3, draw=black!20, line width=0.4pt, rounded corners, inner sep=3.5mm}
}

\begin{scope}[local bounding box=physical_space]
    \node[source_point={\mathbf{p}_1}] (p1) at (0, 1.55) {};
    \node[source_point={\mathbf{p}_2}, right=10mm of p1] (p2) {};
    \node[source_point={\mathbf{p}_3}, right=10mm of p2] (p3) {};
    \node[font=\scriptsize, above=2mm of p2] (source_label) {Source $S$};

    \node[target_point={\mathbf{r}_1}, below=11mm of p1] (r1) {};
    \node[target_point={\mathbf{r}_2}, right=10mm of r1] (r2) {};
    \node[target_point={\mathbf{r}_3}, right=10mm of r2] (r3) {};
    \node[font=\scriptsize, below=2mm of r2] (target_label) {Target $R$};

    \begin{pgfonlayer}{background}
        \node[cloud_container, fit=(p1)(p3)(r1)(r3)(source_label)(target_label)] (clouds_box) {};
    \end{pgfonlayer}

    \draw[association_line] (p1) -- node[pos=0.45, left, font=\scriptsize] {$a_1$} (r1);
    \draw[association_line] (p1) -- node[pos=0.35, right, font=\scriptsize] {$a_2$} (r2);
    \draw[association_line] (p2) -- node[pos=0.70, right, font=\scriptsize] {$a_3$} (r2);
    \draw[association_line] (p2) -- node[pos=0.45, right, font=\scriptsize] {$a_4$} (r3);
    \draw[association_line] (p3) -- node[pos=0.60, right, xshift=1pt, font=\scriptsize] {$a_5$} (r3);

    \node[below=4mm of clouds_box, align=center, font=\scriptsize] (phys_label) {
        Putative associations $a_k=(\mathbf{p}_i,\mathbf{r}_j)\in S\times R$
    };
\end{scope}

\begin{scope}[shift={($(clouds_box.east) + (32mm, -1mm)$)}, local bounding box=graph_space]

    \node[graph_node] (gv1) at (0, 9mm) {$a_1$};
    \node[graph_node, right=13mm of gv1] (gv2) {$a_3$};
    \node[graph_node, below right=6mm and 6mm of gv2] (gv4) {$a_5$};

    \node[graph_node, below=14mm of gv1] (gv5) {$a_2$};
    \node[graph_node, right=8mm of gv5] (gv3) {$a_4$};

    \node[fit=(gv1)(gv2)(gv3)(gv4)(gv5), inner sep=0pt] (graph_content_box) {};

    \draw[consistent_edge] (gv1) -- (gv2);
    \draw[consistent_edge] (gv1) -- (gv4);
    \draw[consistent_edge] (gv2) -- (gv4);
    \draw[consistent_edge] (gv5) -- (gv3);

    \begin{pgfonlayer}{background}
        \node[clique_highlight, fit=(gv1)(gv2)(gv4)] (clique) {};
    \end{pgfonlayer}
    \node[anchor=south west, font=\scriptsize] at (clique.south west) {$C$ (max. clique)};

    \node[align=center, font=\scriptsize, anchor=north] (graph_label) at (graph_content_box.south |- phys_label.north) {
        Consistency graph $G=(V,E)$ over associations.
    };
\end{scope}

\draw[->, draw=black, line width=0.6pt]
    ($(clouds_box.east) + (2mm, 0)$)
    -- node[midway, above, font=\scriptsize] {Construct $G$}
    ($(graph_content_box.west) + (-2mm, 1mm)$);

\end{tikzpicture}

%% file: radar-pcm-invariant.tex
\def\parseXYZ#1,#2,#3;{ \def\valX{#1} \def\valY{#2} \def\valZ{#3} }
\newcommand{\addEllipsoid}[4]{
    \parseXYZ#1; \let\px\valX \let\py\valY \let\pz\valZ % point p
    \parseXYZ#2; \let\spx\valX \let\spy\valY \let\spz\valZ % Sigma_p diagonal
    \parseXYZ#3; \let\sprx\valX \let\spry\valY \let\sprz\valZ % Sigma_p rotation
    % precompute trig
    \pgfmathsetmacro{\cossprz}{cos(\sprz)} 
    \pgfmathsetmacro{\sinsprz}{sin(\sprz)}
    \pgfmathsetmacro{\cosspry}{cos(\spry)} 
    \pgfmathsetmacro{\sinspry}{sin(\spry)}
    \pgfmathsetmacro{\cossprx}{cos(\sprx)} 
    \pgfmathsetmacro{\sinsprx}{sin(\sprx)}
    % speed/quality  
    \pgfmathsetmacro{\EllipsoidSamplesU}{9}
    \pgfmathsetmacro{\EllipsoidSamplesV}{13}

    \addplot3 [
        surf,
        shader=flat,
        z buffer=sort,
        samples=\EllipsoidSamplesU,
        samples y=\EllipsoidSamplesV,
        #4,
        domain=-1:1,     % u
        domain y=0:360,  % v
        declare function={
            %  base elipsoid
            bx(\u,\v) = \spx * sin(\v) * sqrt(1-\u^2);
            by(\u,\v) = \spy * cos(\v) * sqrt(1-\u^2);
            bz(\u,\v) = \spz * \u;
            % rot X (affects y,z)
            rxY(\y,\z) = \y*\cossprx - \z*\sinsprx;
            rxZ(\y,\z) = \y*\sinsprx + \z*\cossprx;
            % rot Y (affects x,z)
            ryX(\x,\z) = \x*\cosspry + \z*\sinspry;
            ryZ(\x,\z) = -\x*\sinspry + \z*\cosspry;
            % rot Z (affects x,y)
            rzX(\x,\y) = \x*\cossprz - \y*\sinsprz;
            rzY(\x,\y) = \x*\sinsprz + \y*\cossprz;
            % order: base -> rot X -> rot Y -> rot Z -> shift
            % nest the functions: ryX uses rxZ's output, rzX uses ryX's output
            finalX(\u,\v) = rzX( ryX(bx(\u,\v), rxZ(by(\u,\v),bz(\u,\v))), rxY(by(\u,\v),bz(\u,\v)) ) + \px;
            finalY(\u,\v) = rzY( ryX(bx(\u,\v), rxZ(by(\u,\v),bz(\u,\v))), rxY(by(\u,\v),bz(\u,\v)) ) + \py;
            finalZ(\u,\v) = ryZ( bx(\u,\v), rxZ(by(\u,\v),bz(\u,\v)) ) + \pz;
        },
    ] 
    (
        { finalX(x,y) },
        { finalY(x,y) },
        { finalZ(x,y) }
    ); 
}

\begin{tikzpicture}
    \tikzset{radarlabel/.style={font=\small, fill=white, fill opacity=0.85, text opacity=1, inner sep=1pt}}

    \def\viewAz{30}
    \def\viewEl{30}

    \begin{axis}
    [%colormap/blackwhite,
    view={\viewAz}{\viewEl},
    axis lines=center,
    axis on top,
    ticks=none,
    set layers=default,
    axis equal,
    clip=false, % no clipping of text floating box pls
    width=\textwidth,
    xmin=0, xmax=13, ymin=0, ymax=13, zmin=0, zmax=10,
    xlabel={$x$}, ylabel={$y$}, zlabel={$z$},
    label style={font=\small},
    xlabel style={anchor=south east},
    ylabel style={anchor=south west},
    zlabel style={anchor=south west},
    enlargelimits=false,
    tick align=inside,
    ]

        \def\pX{  8}   \def\pY{   10}   \def\pZ{  2}
        \def\spX{  0.2}   \def\spY{  2} \def\spZ{1}
        \def\rspX{ 0}   \def\rspY{ 0}   \def\rspZ{50}

        \def\qX{   10}   \def\qY{   2}   \def\qZ{  1}
        \def\sqX{  0.2} \def\sqY{ 2}  \def\sqZ{ 1}
        \def\rsqX{ 0}   \def\rsqY{ 30}   \def\rsqZ{-20}

        \def\rX{  3}   \def\rY{  10}    \def\rZ{  1}
        \def\srX{ 1}  \def\srY{ 1}  \def\srZ{ 0.2}
        \def\rsrX{ 0}   \def\rsrY{ 0}   \def\rsrZ{0}

        \def\sX{   5}   \def\sY{  2}   \def\sZ{  1}
        \def\ssX{  2} \def\ssY{ 2}  \def\ssZ{ 0.2}
        \def\rssX{ 0}   \def\rssY{ 0}   \def\rssZ{00}

        % --- Projected 1D uncertainties along the distance directions (for visualizing $\sigma_v^2$) ---
        % Direction and midpoint for $d_1=\|\mathbf{p}-\mathbf{q}\|$
        \pgfmathsetmacro{\dPQx}{\qX-\pX}
        \pgfmathsetmacro{\dPQy}{\qY-\pY}
        \pgfmathsetmacro{\dPQz}{\qZ-\pZ}
        \pgfmathsetmacro{\lenPQ}{sqrt((\dPQx)^2+(\dPQy)^2+(\dPQz)^2)}
        \pgfmathsetmacro{\uPQx}{\dPQx/\lenPQ}
        \pgfmathsetmacro{\uPQy}{\dPQy/\lenPQ}
        \pgfmathsetmacro{\uPQz}{\dPQz/\lenPQ}
        \pgfmathsetmacro{\mPQx}{(\pX+\qX)/2}
        \pgfmathsetmacro{\mPQy}{(\pY+\qY)/2}
        \pgfmathsetmacro{\mPQz}{(\pZ+\qZ)/2}

        % Direction and midpoint for $d_2=\|\mathbf{r}-\mathbf{s}\|$
        \pgfmathsetmacro{\dRSx}{\sX-\rX}
        \pgfmathsetmacro{\dRSy}{\sY-\rY}
        \pgfmathsetmacro{\dRSz}{\sZ-\rZ}
        \pgfmathsetmacro{\lenRS}{sqrt((\dRSx)^2+(\dRSy)^2+(\dRSz)^2)}
        \pgfmathsetmacro{\uRSx}{\dRSx/\lenRS}
        \pgfmathsetmacro{\uRSy}{\dRSy/\lenRS}
        \pgfmathsetmacro{\uRSz}{\dRSz/\lenRS}
        \pgfmathsetmacro{\mRSx}{(\rX+\sX)/2}
        \pgfmathsetmacro{\mRSy}{(\rY+\sY)/2}
        \pgfmathsetmacro{\mRSz}{(\rZ+\sZ)/2}

        %project a rotated axis-aligned 1$\sigma$ ellipsoid (given by its semi-axes) onto a unit direction.
        % --- For point p (direction along d1) ---
        \pgfmathsetmacro{\uPZX}{\uPQx*cos(-\rspZ) - \uPQy*sin(-\rspZ)}
        \pgfmathsetmacro{\uPZY}{\uPQx*sin(-\rspZ) + \uPQy*cos(-\rspZ)}
        \pgfmathsetmacro{\uPZZ}{\uPQz}
        \pgfmathsetmacro{\uPYX}{\uPZX*cos(-\rspY) + \uPZZ*sin(-\rspY)}
        \pgfmathsetmacro{\uPYY}{\uPZY}
        \pgfmathsetmacro{\uPYZ}{-\uPZX*sin(-\rspY) + \uPZZ*cos(-\rspY)}
        \pgfmathsetmacro{\uPXX}{\uPYX}
        \pgfmathsetmacro{\uPXY}{\uPYY*cos(-\rspX) - \uPYZ*sin(-\rspX)}
        \pgfmathsetmacro{\uPXZ}{\uPYY*sin(-\rspX) + \uPYZ*cos(-\rspX)}
        \pgfmathsetmacro{\sigP}{sqrt((\spX*\uPXX)^2 + (\spY*\uPXY)^2 + (\spZ*\uPXZ)^2)}

        % --- For point q (direction along d1) ---
        \pgfmathsetmacro{\uQZX}{\uPQx*cos(-\rsqZ) - \uPQy*sin(-\rsqZ)}
        \pgfmathsetmacro{\uQZY}{\uPQx*sin(-\rsqZ) + \uPQy*cos(-\rsqZ)}
        \pgfmathsetmacro{\uQZZ}{\uPQz}
        \pgfmathsetmacro{\uQYX}{\uQZX*cos(-\rsqY) + \uQZZ*sin(-\rsqY)}
        \pgfmathsetmacro{\uQYY}{\uQZY}
        \pgfmathsetmacro{\uQYZ}{-\uQZX*sin(-\rsqY) + \uQZZ*cos(-\rsqY)}
        \pgfmathsetmacro{\uQXX}{\uQYX}
        \pgfmathsetmacro{\uQXY}{\uQYY*cos(-\rsqX) - \uQYZ*sin(-\rsqX)}
        \pgfmathsetmacro{\uQXZ}{\uQYY*sin(-\rsqX) + \uQYZ*cos(-\rsqX)}
        \pgfmathsetmacro{\sigQ}{sqrt((\sqX*\uQXX)^2 + (\sqY*\uQXY)^2 + (\sqZ*\uQXZ)^2)}

        \pgfmathsetmacro{\sigDOne}{sqrt((\sigP)^2+(\sigQ)^2)}

        % --- For point r (direction along d2) ---
        \pgfmathsetmacro{\uRZX}{\uRSx*cos(-\rsrZ) - \uRSy*sin(-\rsrZ)}
        \pgfmathsetmacro{\uRZY}{\uRSx*sin(-\rsrZ) + \uRSy*cos(-\rsrZ)}
        \pgfmathsetmacro{\uRZZ}{\uRSz}
        \pgfmathsetmacro{\uRYX}{\uRZX*cos(-\rsrY) + \uRZZ*sin(-\rsrY)}
        \pgfmathsetmacro{\uRYY}{\uRZY}
        \pgfmathsetmacro{\uRYZ}{-\uRZX*sin(-\rsrY) + \uRZZ*cos(-\rsrY)}
        \pgfmathsetmacro{\uRXX}{\uRYX}
        \pgfmathsetmacro{\uRXY}{\uRYY*cos(-\rsrX) - \uRYZ*sin(-\rsrX)}
        \pgfmathsetmacro{\uRXZ}{\uRYY*sin(-\rsrX) + \uRYZ*cos(-\rsrX)}
        \pgfmathsetmacro{\sigR}{sqrt((\srX*\uRXX)^2 + (\srY*\uRXY)^2 + (\srZ*\uRXZ)^2)}

        % --- For point s (direction along d2) ---
        \pgfmathsetmacro{\uSZX}{\uRSx*cos(-\rssZ) - \uRSy*sin(-\rssZ)}
        \pgfmathsetmacro{\uSZY}{\uRSx*sin(-\rssZ) + \uRSy*cos(-\rssZ)}
        \pgfmathsetmacro{\uSZZ}{\uRSz}
        \pgfmathsetmacro{\uSYX}{\uSZX*cos(-\rssY) + \uSZZ*sin(-\rssY)}
        \pgfmathsetmacro{\uSYY}{\uSZY}
        \pgfmathsetmacro{\uSYZ}{-\uSZX*sin(-\rssY) + \uSZZ*cos(-\rssY)}
        \pgfmathsetmacro{\uSXX}{\uSYX}
        \pgfmathsetmacro{\uSXY}{\uSYY*cos(-\rssX) - \uSYZ*sin(-\rssX)}
        \pgfmathsetmacro{\uSXZ}{\uSYY*sin(-\rssX) + \uSYZ*cos(-\rssX)}
        \pgfmathsetmacro{\sigS}{sqrt((\ssX*\uSXX)^2 + (\ssY*\uSXY)^2 + (\ssZ*\uSXZ)^2)}

        \pgfmathsetmacro{\sigDTwo}{sqrt((\sigR)^2+(\sigS)^2)}

        % Variance of v = d1 - d2: sigma_v^2 = sigma_{d1}^2 + sigma_{d2}^2
        \pgfmathsetmacro{\sigV}{sqrt((\sigDOne)^2+(\sigDTwo)^2)}

        % Draw the projected 1D uncertainty segments along the distance directions
        % We offset them in a camera-consistent way: choose an offset direction that is
        % (i) perpendicular to the distance direction and (ii) lies in the image/screen plane.
        % Let \mathbf{c} be the camera viewing direction. Then \mathbf{n}=\mathbf{c}\times\mathbf{u}
        % satisfies \mathbf{n}\perp\mathbf{c} (screen plane) and \mathbf{n}\perp\mathbf{u}.
        \pgfmathsetmacro{\offAmt}{0.35}

        % Approximate camera viewing direction from pgfplots view angles (degrees)
        \pgfmathsetmacro{\camx}{cos(\viewAz)*cos(\viewEl)}
        \pgfmathsetmacro{\camy}{sin(\viewAz)*cos(\viewEl)}
        \pgfmathsetmacro{\camz}{sin(\viewEl)}

        % Offset direction for d1: nPQ = cam x uPQ
        \pgfmathsetmacro{\nPQx}{\camy*\uPQz-\camz*\uPQy}
        \pgfmathsetmacro{\nPQy}{\camz*\uPQx-\camx*\uPQz}
        \pgfmathsetmacro{\nPQz}{\camx*\uPQy-\camy*\uPQx}
        \pgfmathsetmacro{\nPQlen}{sqrt((\nPQx)^2+(\nPQy)^2+(\nPQz)^2)}
        \pgfmathsetmacro{\nPQxU}{\nPQx/\nPQlen}
        \pgfmathsetmacro{\nPQyU}{\nPQy/\nPQlen}
        \pgfmathsetmacro{\nPQzU}{\nPQz/\nPQlen}

        % Offset direction for d2: nRS = cam x uRS
        \pgfmathsetmacro{\nRSx}{\camy*\uRSz-\camz*\uRSy}
        \pgfmathsetmacro{\nRSy}{\camz*\uRSx-\camx*\uRSz}
        \pgfmathsetmacro{\nRSz}{\camx*\uRSy-\camy*\uRSx}
        \pgfmathsetmacro{\nRSlen}{sqrt((\nRSx)^2+(\nRSy)^2+(\nRSz)^2)}
        \pgfmathsetmacro{\nRSxU}{\nRSx/\nRSlen}
        \pgfmathsetmacro{\nRSyU}{\nRSy/\nRSlen}
        \pgfmathsetmacro{\nRSzU}{\nRSz/\nRSlen}

        \draw[blue!60!black, very thick]
            (axis cs:{\mPQx-\uPQx*\sigDOne+\nPQxU*\offAmt},{\mPQy-\uPQy*\sigDOne+\nPQyU*\offAmt},{\mPQz-\uPQz*\sigDOne+\nPQzU*\offAmt})
            -- node[radarlabel, midway, above, xshift=10pt] {$\pm\,\sigma_{d_1}$}
            (axis cs:{\mPQx+\uPQx*\sigDOne+\nPQxU*\offAmt},{\mPQy+\uPQy*\sigDOne+\nPQyU*\offAmt},{\mPQz+\uPQz*\sigDOne+\nPQzU*\offAmt});

        \draw[orange!70!black, very thick]
            (axis cs:{\mRSx-\uRSx*\sigDTwo+\nRSxU*\offAmt},{\mRSy-\uRSy*\sigDTwo+\nRSyU*\offAmt},{\mRSz-\uRSz*\sigDTwo+\nRSzU*\offAmt})
            -- node[radarlabel, midway, above, xshift=10pt] {$\pm\,\sigma_{d_2}$}
            (axis cs:{\mRSx+\uRSx*\sigDTwo+\nRSxU*\offAmt},{\mRSy+\uRSy*\sigDTwo+\nRSyU*\offAmt},{\mRSz+\uRSz*\sigDTwo+\nRSzU*\offAmt});

        \node[radarlabel, anchor=center] at (axis description cs:0.5,-0.10) {\begin{tikzpicture}[baseline=(current bounding box.center), font=\footnotesize]
            \pgfmathsetlengthmacro{\Lone}{\sigDOne*1.00cm}
            \pgfmathsetlengthmacro{\Ltwo}{\sigDTwo*1.00cm}
            \pgfmathsetlengthmacro{\Lv}{\sigV*1.00cm}
            \pgfmathsetlengthmacro{\Lmax}{max(max(\Lone,\Ltwo),\Lv)}

            % Layout: use a TikZ matrix so label baselines are aligned and never collide/shift.
            \matrix (bars) [
                matrix of nodes,
                ampersand replacement=\&,
                nodes={inner sep=0pt, outer sep=0pt, anchor=west},
                %row sep=10pt,
                %column sep=10pt
            ] {
                $\sigma_{d_1}$ \& |[coordinate] (b1)| \\
                $\sigma_{d_2}$ \& |[coordinate] (b2)| \\
                $\sigma_{v}$   \& |[coordinate] (b3)| \\
            };

            % sigma bars
            \coordinate (bar0) at ([xshift=8pt]b1);

            \draw[black!40, line width=0.8pt] (bar0 |- b3) ++(0,-4pt) -- ++(\Lmax,0);
            \node[anchor=north, inner sep=0pt, font=\scriptsize] at ($(bar0 |- b3)+(0,-4pt)$) {0};

            \draw[blue!60!black, line width=2.2pt] (bar0 |- b1) -- ++(\Lone,0);
            \draw[orange!70!black, line width=2.2pt] (bar0 |- b2) -- ++(\Ltwo,0);
            \draw[black, line width=2.2pt] (bar0 |- b3) -- ++(\Lv,0);
        \end{tikzpicture}};

        \draw plot (\pX, \pY, \pZ) node [radarlabel, above right, xshift=10] {$\mathbf{p}, \boldsymbol{\Sigma}_{\mathbf{p}}$};
        \draw plot (\qX, \qY, \qZ) node [radarlabel, below right, xshift=15] {$\mathbf{q}, \boldsymbol{\Sigma}_{\mathbf{q}}$};
        \draw plot (\rX, \rY, \rZ) node [radarlabel, left, xshift=-20] {$\mathbf{r}, \boldsymbol{\Sigma}_{\mathbf{r}}$};
        \draw plot (\sX, \sY, \sZ) node [radarlabel, below left, xshift=-40pt] {$\mathbf{s}, \boldsymbol{\Sigma}_{\mathbf{s}}$};

        \draw[black, thick, dashed]
        (\pX, \pY, \pZ)
            -- node[radarlabel, midway, left] {$d_1$}
        (\qX, \qY, \qZ);
        \draw[black, thick, dashed]
        (\rX, \rY, \rZ)
            -- node[radarlabel, midway, left] {$d_2$}
        (\sX, \sY, \sZ);

        \addplot3 [
        only marks,
        mark=*,
        mark size=1.2pt,
        color=black
        ] coordinates {
            (\pX, \pY, \pZ)
            (\qX, \qY, \qZ)
            (\rX, \rY, \rZ)
            (\sX, \sY, \sZ)
        };

        \addEllipsoid{\pX,\pY,\pZ}{\spX,\spY,\spZ}{\rspX,\rspY,\rspZ}{draw=blue!60!black, fill=blue, fill opacity=0.2}
        \addEllipsoid{\qX,\qY,\qZ}{\sqX,\sqY,\sqZ}{\rsqX,\rsqY,\rsqZ}{draw=blue!60!black, fill=blue, fill opacity=0.2}
        \addEllipsoid{\rX,\rY,\rZ}{\srX,\srY,\srZ}{\rsrX,\rsrY,\rsrZ}{draw=orange!70!black, fill=orange, fill opacity=0.2}
        \addEllipsoid{\sX,\sY,\sZ}{\ssX,\ssY,\ssZ}{\rssX,\rssY,\rssZ}{draw=orange!70!black, fill=orange, fill opacity=0.2}
        
    \end{axis}
\end{tikzpicture}

%% file: icp-pcm.tex
\begin{tikzpicture}[
    >=Latex,
    font=\sffamily,
    node distance=12mm and 11mm,
    block/.style={rectangle, draw, thick, fill=blue!5, text width=3.8cm, text centered, rounded corners, minimum height=11mm},
    cloud/.style={ellipse, draw, thick, fill=gray!10, text width=2.8cm, text centered, minimum height=10mm, inner sep=1mm},
    decision/.style={diamond, draw, thick, fill=yellow!10, text width=18mm, text badly centered, inner sep=0pt},
    data/.style={trapezium, trapezium left angle=70, trapezium right angle=110, draw, thick, fill=green!10, text width=2.6cm, text centered, minimum height=10mm, inner sep=1mm},
    line/.style={draw, thick, ->},
    loop/.style={draw, thick, ->, rounded corners=2pt},
    pcmblock/.style={block, fill=orange!15, draw=orange!70!black, very thick}
]

\node[block] (assoc) {Generate correspondences\\$\mathcal{A}_k$};
\node[pcmblock, right=of assoc] (pcm) {\ac{PCM} inlier selection\\$\mathcal{I}_k\subset\mathcal{A}_k$};
\node[block, right=of pcm] (solve) {Solve \ac{ICP} step\\$\Delta\mathbf{T}_k$};

\node[cloud, left=of assoc] (src) {Source cloud\\$S$};
\node[cloud, below=of src] (tgt) {Target cloud\\$R$};

\node[block, below=of solve] (update) {Update transform\\$\mathbf{T}_{k+1}=\Delta\mathbf{T}_k\,\mathbf{T}_k$};
\node[decision, left=of update] (check) {Stop?};
\node[data, left=of check] (out) {Output\\$\mathbf{T}$};

\coordinate (assocInTop) at ([yshift=3mm]assoc.west);
\coordinate (assocInBot) at ([yshift=-3mm]assoc.west);

\draw[line] (src.east) |- (assocInTop);
\draw[line] (tgt.east) |- (assocInBot);
\draw[line] (assoc) -- (pcm) node[midway, above] {$\mathcal{A}_k$};
\draw[line] (pcm) -- (solve) node[midway, above] {$\mathcal{I}_k$};
\draw[line] (solve.south) -- (update.north) node[midway, right] {$\Delta\mathbf{T}_k$};
\draw[line] (update.west) -- (check.east);
\draw[line] (check.west) -- (out.east) node[midway, above] {yes};

\draw[line] (check.north) |- ([yshift=3mm]check.north) -| (assoc.south) node[pos=0.25, below] {no};

\end{tikzpicture}